\documentclass{article}
\usepackage{ijcai22}

\usepackage{times}
\usepackage{soul}
\usepackage{url}
\usepackage{xr}
\usepackage[hidelinks]{hyperref}
\usepackage[utf8]{inputenc}
\usepackage[small]{caption}
\usepackage{graphicx}
\usepackage{amsmath}
\usepackage{amsthm} 
\usepackage{booktabs}
\usepackage{amsmath,amssymb}
\usepackage{algorithm}
\usepackage{algpseudocode}
\usepackage{siunitx}
\usepackage{color}
\usepackage{xcolor}
\newcommand{\minisection}[1]{\vspace{0.03in} \noindent {\bf #1}}

\usepackage{todonotes}
\usepackage{natbib}

\DeclareMathOperator{\EX}{\mathbb{E}}

\def\BibTeX{{\rm B\kern-.05em{\sc i\kern-.025em b}\kern-.08em
    T\kern-.1667em\lower.7ex\hbox{E}\kern-.125emX}}

\usepackage[skip=6pt]{caption}

\title{CCPT: Automatic Gameplay Testing and Validation with Curiosity-Conditioned Proximal Trajectories}

\author{
Alessandro Sestini$^{1,2}$
\and
Linus Gisslén$^1$\and
Joakim Bergdahl$^1$\and
Konrad Tollmar$^1$\And
Andrew D. Bagdanov$^2$
\affiliations
$^1$SEED - Electronic Arts (EA)\\
$^2$Università degli Studi di Firenze
\emails
\{asestini, lgisslen, jbergdahl, ktollmar\}@ea.com \\
andrew.bagdanov@unifi.it
}

\begin{document}

\maketitle

\begin{abstract}

This paper proposes a novel deep reinforcement learning algorithm to perform automatic analysis and detection of gameplay issues in complex 3D navigation environments. The Curiosity-Conditioned Proximal Trajectories (CCPT) method combines curiosity and imitation learning to train agents to methodically explore in the proximity of known trajectories derived from expert demonstrations. We show how CCPT can explore complex environments, discover gameplay issues and design oversights in the process, and recognize and highlight them directly to game designers. We further demonstrate the effectiveness of the algorithm in a novel 3D navigation environment which reflects the complexity of modern AAA video games. Our results show a higher level of coverage and bug discovery than baselines methods, and it hence can provide a valuable tool for game designers to identify issues in game design automatically.
\end{abstract}

\section{Introduction}
\label{sec:intro}
Play testing plays a crucial role in the production of modern video games. The presence of gameplay issues and bugs can greatly decrease the overall player experience and therefore is crucial to be kept to a minimum. However, modern video games have grown both in size and complexity and
thorough coverage is not often feasible via manual human play testing. 
The goal of automated gameplay testing is to free some of the human resources to do more ``meaningful'' testing such as measuring gameplay balance, difficulty, and potential retention rate.

Recently, automated testing approaches have been proposed to mitigate total reliance on human testers by training AI-based agents to explore
large game scenes~\citep{improving}. Automatic exploration through
intrinsic motivation is a step in the right direction,
however we argue that it is not enough.
First, we need agents capable of learning the world around them, 
efficiently understanding the difference between different states.
Second, we need agents that know the 
difference between good trajectories and bad trajectories, in order to 
recognize which paths are ``broken'' ones.  

\begin{figure}
    \centering \includegraphics[width=0.95\columnwidth]{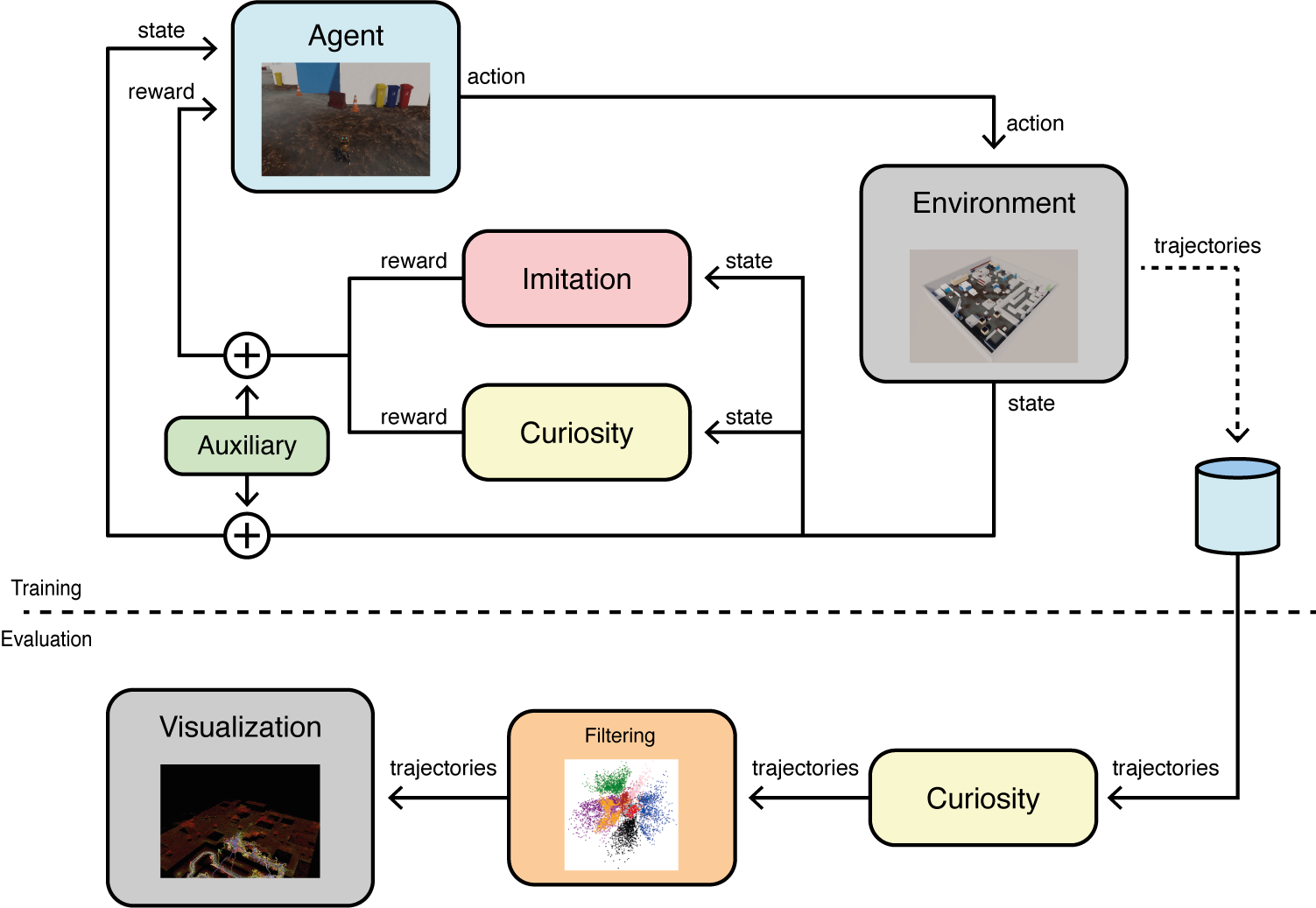}
    \caption{
    Overview of our approach. By combining imitation learning and curiosity, 
    we train agents to play test a large game scenario.
    All information gathered during training is saved
    and during evaluation is filtered
    through the same curiosity module used during training.
    Our \emph{exploration-conditioned intrinsic reward
    function} enables us to filter and highlight
    trajectories that contain bugs and design oversights
    like missing collision boxes or gameplay glitches.} 
    \label{fig:teasing}
\end{figure}

Therefore, in this paper we propose a novel reinforcement learning (RL) approach able to train agents which can explore and analyze a large 3D environment composed of complex navigation challenges. We call the approach Automatic Gameplay Testing and Validation with Curiosity-Conditioned Proximal Trajectories (CCPT). As shown in Figure~\ref{fig:teasing}, our
technique use a combination of \textit{curiosity}, which leads agents to seek novel interactions and to improve exploration,
and \textit{imitation learning} which lets agents explore the proximity of demonstrated trajectories. In particular, we propose an \textit{exploration-conditioned
intrinsic reward function} leading to agents that do not just
learn to be curious, but that learn what it means to behave
curiously. In fact, we train agents which we can control
to behave ``target-driven'' versus ``exploration-driven".
The model enables us not only to find bugs and
issues, but to automatically identify, filter and highlight them among the massive quantity of information collected by agents during their interactions with the environment. 

    
    
    

Our key contributions are:
1) a new open source environment for complex 3D
    navigation challenges, suitable as testbed for training both
    exploratory and navigation agents;
2) a novel neural network architecture for navigation
    and exploration agents and an empirical demonstration of its 
    effectiveness;
3) a new exploration algorithm which, thanks to
    its use of expert demonstrations, is able to deliberately explore the proximity
    of desired trajectories; and
4) our results show how these approaches can help video game
    designers automatically detect bugs and issues in complex 3D game scenarios.


\section{Related Work}
\label{sec:related}
The potential of deep reinforcement learning for video game testing has been gaining interest
from both the research and video game communities. Here we review
recent work most related to our contributions.

\minisection{Automated Play Testing.} Several recent studies
have investigated the use of AI techniques to perform automatic play testing,
with a focus on maximizing game state coverage. Many of these recent works heavily rely on classical hand-scripted AI or random exploration 
\citep{pathos,personas}. However, when dealing with complex 3D environments with difficult navigation challenges we argue
that these techniques are not readily applicable due to the
high-dimensional state-space. The Reveal-More algorithm
also uses human demonstrations to guide the random exploration,
although in simple 2D dungeon levels~\citep{reveal}. \cite{mugrai2019automated} developed an algorithm to mimic human behaviour to get more meaningful gameplay testing, but also to aid in the game design.

At the same time, many other works have used reinforcement learning to perform
either automatic play testing or complex navigation exploration. 
\citet{ubisoftnavigation} trained a reinforcement learning agent to navigate a complex
3D environment toward procedurally-generated goals, while 
\cite{microsoft_navigation} trained different agents to perform a Turing
test to evaluate the human-likeness of trained bots.  Closer
to our work, \cite{2dplaytest} trained reinforcement learning agents to perform
automated play testing in 2D side-scrolling games, also providing a set
of visualizations for level design analysis, and \cite{improving} used
intrinsic motivation to train many agents to explore a 3D scenario
with the aim of finding issues and oversights.

Although our work draws inspiration from \cite{improving}, they based their approach on count-based exploration that may not be feasible when faced with complex environmental dynamics that, due to the tabular nature of such algorithms, explode in complexity~\citep{count}. Moreover, the use of a purely
exploration-based technique can slow down coverage time, especially if designers want to test a particular part of the environment. Finally, even given good visualizations of the results, this approach does not tell where, when, and how the issues are found, but rather leave the burden of recognizing them to the designers.

\minisection{Imitation Learning.} Similar to \cite{reveal}, we make use of demonstrations to guide the exploration. However, for this aim we use a state-of-the-art imitation learning algorithm.  Imitation learning aims to distill a policy mimicking the behavior of an expert demonstrator from a dataset of demonstrations. It is often assumed that demonstrations come from an expert who is behaving near-optimally. Standard approaches are based on Behavioral Cloning that mainly use supervised
learning~\citep{bc1,dagger,tamer}. Generative Adversarial 
Imitation Learning (GAIL) is a recent imitation learning technique which is based on a generator-discriminator approach~\citep{gail}, as \cite{connection} noticed that imitation learning is closely related to the training of Generative Adversarial Networks (GAN). Based on ideas from GAIL, \cite{amp} proposed the Adversarial Motion Prior (AMP) algorithm, which is a GAN-based imitation learning method 
that aims to increase stability of adversarial approaches.

\minisection{Intrinsic motivation.} Intrinsic motivation aims to encourage agents to explore the environment states in the absence of an extrinsic reward. The already mentioned count-based exploration is a natural way to do exploration, although for high-dimensional state spaces it can be infeasible~\citep{count}. Another class of exploration methods rely on errors in predicting dynamics~\citep{curiosity,rnd}. These are machine learning techniques for high-dimensional states that aim to push agents to explore never or less-encountered states during training. The interested reader should consult~\cite{imsurvey} for a detailed survey of the state-of-the-art in intrinsic motivation.

\section{Proposed Method}
\label{sec:method}
This section details our approach to train agents guided by expert priors to find bugs and gameplay issues. 

\subsection{The Navigation Environment}
\label{sec:env}
\begin{figure}
    \begin{center}
    \scalebox{0.95}{
    \begin{tabular}{cc}
        \includegraphics[width=0.285\textwidth]{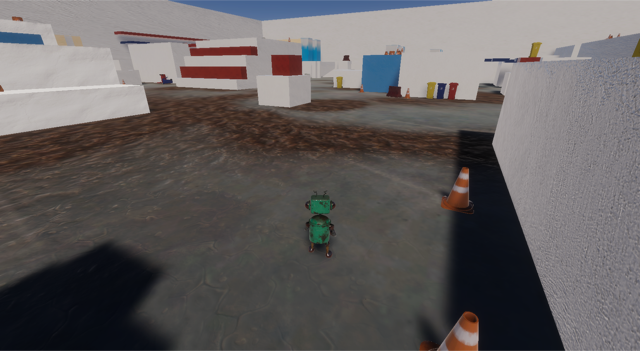} &
        \includegraphics[width=0.155\textwidth]{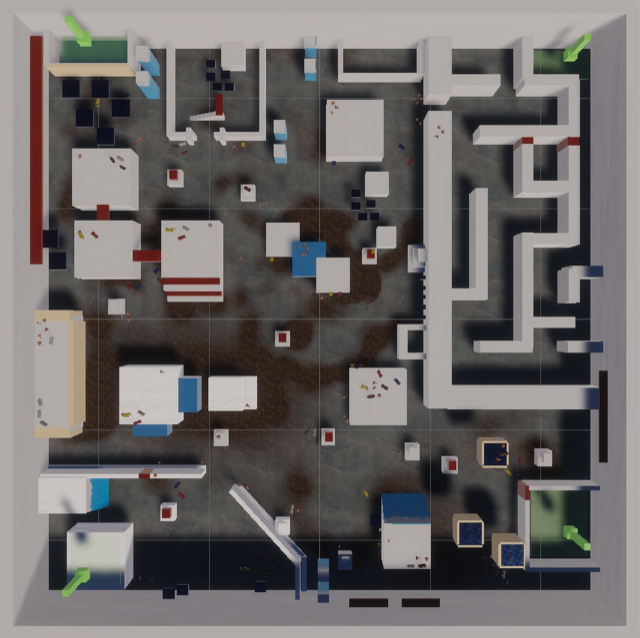}  \\
        (a) Screenshot of the environment. & (b) Top-down view.
    \end{tabular}
    }
    \end{center}
    \caption{Overview of our proposed environment.}
    \label{fig:env_screen}
\end{figure}

To validate our approach, and to support continued research, we propose a 3D navigation environment designed
to resemble modern AAA game scenarios. A screenshot of the environment is given in Figure~\ref{fig:env_screen} together with a top-down view of the whole map. The scene is approximately $\SI{500}{\m}\times \SI{500}{\m} \times \SI{60}{\m}$ in size and contains a variety of navigation challenges and dynamic elements such as moving platforms and elevators. Agents wishing to explore all secrets contained in the map must learn complex navigation strategies. Our
environment is comparable to recent navigation studies
\citep{improving,microsoft_navigation,ubisoftnavigation} and is open source and available for anyone who wants to exploit, contribute, or expand on this research.\footnote{Repository to be published upon acceptance.} For more details about the navigation mechanics, see the accompanying video included in the Supplementary Material.

Agents spawn in the center of the map and each episode
ends after 500 timesteps. There are four goal areas in the environment selected as they are the most difficult spots to reach (indicated by the green arrows in Figure~\ref{fig:env_screen}(b)). The
agent has a total of 10 discrete actions: move
forward/backward/left/right, move in one of the 4 diagonal directions,
jump, and wait. The agent can also perform a double
jump while in the air and climb on surfaces of specific elements
located around the map.
Moreover, since the intent of the work is to provide an automated way
to detect bugs and glitches in a game scene, we manually introduce
such gameplay issues like missing collision
boxes and glitches throughout the map.

The state observed by the agent at any instant in time is composed of a local
perception in the form of a 3D semantic occupancy map, as well as
scalar information about physical attributes of the agent (global
position, if is grounded, if is attached to a climbable surface, if it
can perform a double jump,  its velocity and direction). We show an example 3D
semantic occupancy map as inputs to the networks in Figure~\ref{fig:nets}. 
These maps are a categorical discretization of the space and elements around the agent,
and each voxel is defined by the semantic integer value of the type of
object at that position. 

The only extrinsic reward provided is given when an
agent reaches an active goal, for which it receives +10 for
each timestep it stays inside this area.
With such a sparse reward in such a large environment, agents
have very little change of receiving even a single, non-zero reward, thus making training very hard.
Moreover, with just the extrinsic reward, even if they learn
to reach a goal location, agents will always follow the same trajectory without exploring for new paths.
Instead we need agents able to efficiently arrive to a goal area while continuing to search for undiscovered paths, 
thus combining rewards for both exploration and imitation.


\subsection{The CCPT Algorithm}
\label{sec:alg}
In this section we detail the algorithm used to generate agents to perform automated play testing. We devise our method following a main idea: train agents which explore in the \emph{proximity} of trajectories predefined by an expert in order to automatically identify overlooked issues.
We define agents composed of navigation, imitation, exploration sub-modules which contribute to the reward function used to drive policy learning.

\begin{figure}
    \begin{center}
    \scalebox{0.88}{
    \begin{tabular}{c}
        \includegraphics[width=\columnwidth]{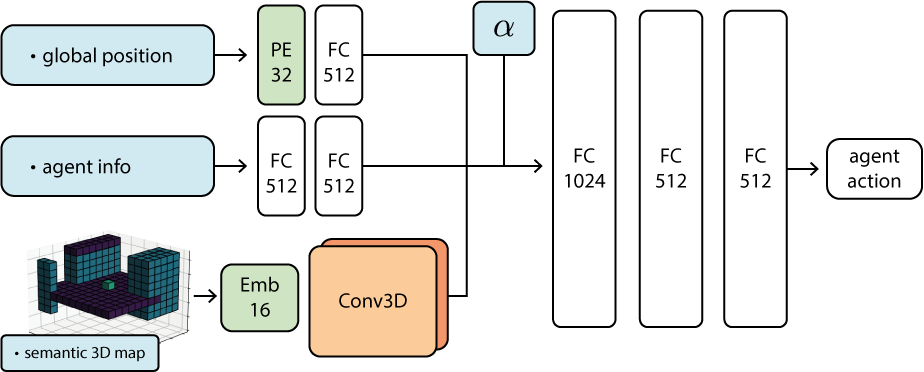} \\
        (a) Navigation module \\
        \includegraphics[width=\columnwidth]{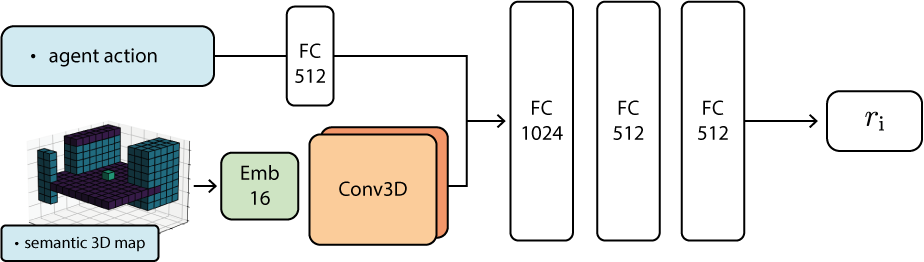} \\
        (b) Imitation module \\
        \includegraphics[width=\columnwidth]{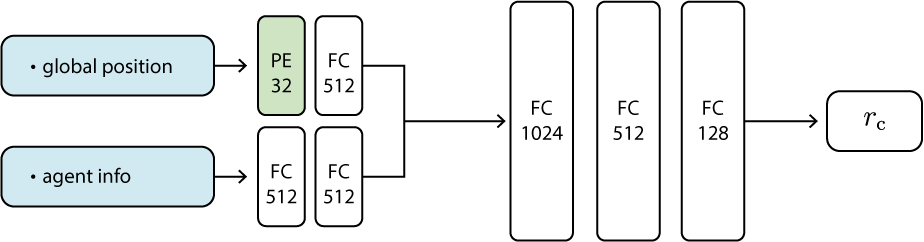} \\
        (c) Curiosity module
    \end{tabular}
    }
    \end{center}
    \caption{Overview of the module architectures used in this work. Additional details are given in the Supplementary Material.}
    \label{fig:nets}
\end{figure}

\minisection{Navigation Module.}
The navigation module is defined by the policy. As shown in
Figure \ref{fig:nets}(a), the policy takes as input both the global information and the local perception defined by the
semantic occupancy map. It also takes an auxiliary input
that defines the level of exploration followed in a particular
episode. Since it is a fundamental part of the reward function, we defer the description of how this affects training to the section Reward Function below. To encode the
global 3D position $(X, Y, Z)$ of the agent we use positional
embeddings of~\cite{attention}:
\begin{equation}
\label{eq:pos_embs}
    \begin{gathered}
        \text{PE}_{(\text{pos}, 2i)} = \sin{\frac{\text{pos}}{10000^{\frac{2i}{\text{d}}}}}
        \;\\
        \text{PE}_{(\text{pos}, 2i + 1)} = \cos{\frac{\text{pos}}{10000^{\frac{2i}{\text{d}}}}},
    \end{gathered}
\end{equation}
where $pos$ is the integer position, $i \in \{0, 1, 2\}$ is the coordinate being encoded, and $d$ is 
the embedding size. 
We claim that the use of such an embedding is crucial
for training agents that navigate and explore these environments.
In fact, the general way of encoding such information with
normalization or with learned embeddings it is not enough
to efficiently understand the difference between different
states. In Section \ref{sec:exp} we support this claim
with ablation experiments.

These vectors are concatenated with the other agent information and passed through a feedforward network. The semantic occupancy map is instead passed through its own 3D convolutional network. All the processed vectors are then concatenated together and passed through an
MLP. The policy is trained using the Proximal Policy Optimization (PPO) algorithm~\citep{ppo}.

\minisection{Imitation Module.}
The imitation module trains the agent to follow the expert
trajectories and to guide play testing toward a particular
area.  We use the AMP algorithm~\citep{amp}, which is built on top of GAIL~\citep{gail}.  Given a set of expert
demonstrations $M$, the goal is to learn to measure the
similarity between the policy and the demonstrations, and to then update the policy via forward-RL. The objective is modeled as a discriminator $D(s, a)$ trained to predict whether a given state-action pair $(s,a)$ is
sampled from the demonstration set or generated by running the
policy. AMP adopts the loss function proposed for
the least-square GAN~\citep{leastgan}:
\begin{align}
    \mathcal{L}^{\text{AMP}} = \text{arg min}_D & \EX_{d^M(s,a)} [(D(s,a)-1)^2] + \nonumber
    \\ & \EX_{d^\pi(s,a)}[(D(s,a)+1)^2],
\end{align}
where $d^M(s,a)$ and $d^\pi(s,a)$ respectively denote the likelihood of observing a
state-action pair in the dataset $M$ or by following the policy $\pi$. The reward function for training the policy is then given by:
\begin{equation}
    r_{\text{i}}(s_t, a_t) = \max{[0, 1 - 0.25 (D(s_t, a_t) - 1)^2]}.
\end{equation}
To further increase training stability, we apply gradient
penalties that penalize nonzero gradients on samples from the
dataset~\citep{gradients}.

The state $s$ of the imitation module is described by the local perception of the agent, which is defined by the 3D semantic occupancy map. This is passed through a 3D convolutional network and is
concatenated with the action embedding before being fed to a feedforward network, as shown in Figure~\ref{fig:nets}(b).

\minisection{Curiosity Module.}
The curiosity module is responsible for optimizing coverage of the environment in the neighborhood of expert demonstrations via intrinsic exploration. Instead of using count-based exploration like~\cite{improving}, which
can be infeasible for high-dimensional state spaces, we use the Random Network Distillation (RND) algorithm~\citep{rnd}. The curiosity module gives an intrinsic reward that is higher for novel and less-encountered states. With this
reward we can train agents with forward-RL to increase coverage of the environment.

The RND algorithm uses two neural networks: a fixed and
randomly initialized target network $\hat{\phi}$ which establishes the prediction problem, and a predictor network $\phi$ trained on data collected by the agent. The predictor network is trained by gradient descent to minimize the expected MSE:
\begin{equation}
    \mathcal{L}^{\text{RND}} = (\hat{\phi}(s) - \phi(s))^2,
\end{equation}
with respect to its parameters. This process distills a randomly initialized neural network into a trained one. The reward $r_{\text{c}}$ of the curiosity module is the same MSE used to train the network:
\begin{equation}
    r_{\text{c}}(s_t) = (\hat{\phi}(s_{t+1}) - \phi(s_{t+1}))^2.
\end{equation}
The more a state is visited by agents, the closer the output of the predictor network will be to that of the target network for that particular state, lowering the prediction error and thus the reward signal for exploration. States encountered following the expert demonstrations will produce low reward values, while for states encountered less frequently the predictor will not be able to perfectly replicate the target, increasing the reward signal and guiding agents toward undiscovered paths.

As shown in Figure \ref{fig:nets}(c), both target and predictor networks take as input the global position, encoded with the positional embedding of Equation~\ref{eq:pos_embs}, and the other agent information. These vectors are then concatenated and passed through a feedforward network. 

\minisection{Reward Function.}
The core of the algorithm lies in the reward function. Our aim is to combine the above modules to derive agents
that can explore the proximity of expert trajectories. In this
work we propose an \textit{exploration-conditioned intrinsic
reward function}. Inspired by works like \citep{ubisoft_conditioned} and \citep{ea_adversarial}, or in general by goal-conditioned
policies~\citep{her}, our reward function is:
\begin{align}
\label{eq:reward}
    R(s_t, a_t) = & \; \alpha \cdot r_{\text{c}}(s_{t+1}) + (1 - \alpha) \cdot r_{\text{i}}(s_t, a_t) \nonumber
    \\
    & + r_{\text{e}}(s_t, a_t),
\end{align}
where $r_{\text{c}}$ is the reward from the curiosity module,
$r_{\text{i}}$ is the reward from the imitation module,
$r_{\text{e}}$ is the extrinsic reward from the environment, and $\alpha \in [0, 1]$ is a weight hyperparameter that controls the level of exploration versus imitation. The value of $\alpha$ is randomly sampled at the beginning of each episode and remains fixed for all timesteps. If an agent samples a value of $\alpha < 0.5$, the
$r_{\text{i}}$ prevails and the reward leads agents to follow the expert demonstrations more closely.  When $\alpha = 0$, the reward moves agents toward a perfect replication of expert trajectories. In contrast, if $\alpha > 0.5$ the $r_{\text{c}}$ is dominant and the reward lead the agent to explore more.  The greater the
$\alpha$, the farther away from expert priors agents explore. When $\alpha = 1$, the agent completely avoids the expert
demonstrations, finding completely new ways to arrive to the goal location. The $r_{\text{e}}$ is needed to make agents arrive in the goal area independently of the sampled $\alpha$.

In order for the agent to understand in which way it should behave in a particular episode, the sampled $\alpha$ is part of the state space of the agent. Since $\alpha$ controls the reward that the agent receives, we are basically combining curiosity-driven and goal-conditioned reinforcement learning. In this setting we get a meaningful exploration via curiosity and not just timestep-level randomness, and the result of training is not just a curious agent, but an agent which we can control to behave like the expert or like an explorer just by changing the $\alpha$ in input to the agent. 



\subsection{Highlighting Suspicious Trajectories}
\label{sec:filter}
The final result of our algorithm is not trained agents, but rather all the information gathered during training. Given the set of all trajectories performed during training, our aim is to find those that evidence game behavior unintended by designers. Since in our case the intended paths are described by the demonstrations, we must find trajectories that arrive at the same goal area defined by experts but are very different from the expected experience.

Thanks to the $\alpha$ component of the reward function in Equation \ref{eq:reward}, for low
$\alpha$ values the agent will revert to following the expert
demonstrations very closely, thus lowering the rewards output by the curiosity module for states that are very near to those seen in expert demonstrations. In contrast, as $\alpha$ increases agents will explore more and more, and the rewards output by the curiosity module for states explored in this setting will be kept relatively high with respect to those near to the expert demonstrations.

We perform a simple first filtering of trajectories by removing all those gathered with $\alpha < 0.5$.  For the the remaining trajectories we exploit the values of the curiosity
module. Given all trajectories that arrive at the goal
location with $\alpha \geq 0.5$, we compute the average curiosity values along the trajectory from the start to the goal location:
\begin{equation}
    \hat{r}_{\text{c}}(\theta_i) = \frac{\sum_{t=0}^{T}{r_{\text{c}}(s_t)}}{T},
\end{equation}
where $\theta_i = (s^i_0, a^i_0, ..., s^i_T, a^i_T)$ is a trajectory, $T$ is
the number of timesteps to arrive to the goal location, and
$r_{\text{c}}$ is the reward of the curiosity module at the end of the
training.  We then define the set $\Theta$ as:
\begin{equation}
    \Theta = \{\theta_i \: | \: \hat{r}_{\text{c}}(\theta_i) > \epsilon\},
\end{equation}
where $\epsilon$ is a pre-defined threshold. As we will show in Section \ref{sec:exp}, this set will define the trajectories that are far from the expert demonstrations and probably display unintended game behavior. Algorithm \ref{alg:alg} details the full
training procedure of CCPT.

\begin{algorithm}[t]
\scriptsize
\caption{Training with CCPT}\label{alg:alg}
\begin{algorithmic}
\State \textbf{input: }$M$ dataset of expert demonstrations, $\epsilon$ filter threshold
\State $\pi \leftarrow$ initialize policy
\State $D \leftarrow$ initialize AMP discriminator
\State $\hat{\phi} \leftarrow$ initialize RND target network
\State $\phi \leftarrow$ initialize RND predictor network
\State $G \leftarrow$ initialize external dataset
\medskip
\While{not done} \Comment{models training}

    \For{$i= 1, ..., m$}
        \State $B \leftarrow$ initialize policy experience dataset
        \State $\alpha_i \sim [0, 1]$ \Comment{sample exploration value}
        \State $\theta_i \sim \pi$ \Comment{sample trajectory} 
        \For{$t=1, ..., T$}
            \State $r^t_{\text{i}} = \max{[0, 1 - 0.25 (D(s_t, a_t) - 1)^2]}$
            \State $r^t_{\text{c}} = ||(\hat{\phi}(s_{t+1}) - \phi(s_{t+1}))^2 ||$
            \State $R^t = \alpha_i \cdot r^t_{\text{c}} + (1 - \alpha_i) \cdot r^t_{\text{i}} + r^t_{\text{e}}$
            \State Store $R^t$ in $\theta_i$
        \EndFor
    \State Store $\theta_i$ in $B$
    \State Store ($\theta_i, \alpha_i$) in $G$ \Comment{store all trajectories in the external dataset}
    \EndFor
    \State Update $D$ with $\mathcal{L}^{\text{AMP}}$ with samples from $B$
    \State Update $\phi$ with $\mathcal{L}^{\text{RND}}$ with samples from $B$
    \State Update $\pi$ with $\mathcal{L}^{\text{PPO}}$ with samples from $B$
\EndWhile
\medskip
\State $\Theta \leftarrow$ initialize set
\For{trajectory $\theta_i \in G \: | \: \alpha_i >= 0.5$} \Comment{filtering results}
    \State $\hat{r}^i_{\text{c}} = \frac{\sum_{t=0}^{T}{r_{\text{c}}(s_t)}}{T}$
    \If{$\hat{r}^i_{\text{c}}(\theta_i) > \epsilon$}
        \State Store $\theta_i$ in $\Theta$
    \EndIf
    
\EndFor
\State \Return{$\Theta$} \Comment{return set of broken trajectories}

\end{algorithmic}
\end{algorithm}

\section{Experimental Results}
\label{sec:exp}
In this section we detail experiments that showcase the capability of
our algorithm to perform automated play testing. We are interested
in three primary research questions:
1) Can CCPT find and highlight bugs and oversights in a 
    complex 3D environment?
2) How does the \textit{exploration-conditioned intrinsic reward function} improve play testing efficiency?
3) Do our models offer better performance than traditional
    models used for 3D navigation?
For all experiments in this section we used our navigation environment described in Section \ref{sec:env}. We report on additional experiments using the VizDoom 
environment \citep{vizdoom} and give complete details on hyperparameters used in all experiments in the Supplementary Material.


\subsection{Play Testing Performance}
To evaluate the ability of CCPT to find and highlight
bugs we tested four different goal areas in the environment. These areas represent four of the most difficult spots to
reach, with trajectories that include dynamic elements, climbable surfaces and complex navigation challenges. For the goal areas we provide six expert demonstrations for each one showing the intended way to arrive
at each specific goal. We then train ten agents 
in parallel and show the most relevant trajectories found by the algorithm for each area.


Table \ref{tab:results} summarizes the results found for all four areas. CCPT is able to find and highlight different ways of arriving to the same goal area of the expert demonstrations, however taking different paths and using different elements with respect to the intended ones. Compared to the baselines, our CCPT clearly outperforms other methods not only in finding bugs we manually inserted in the environment, but also in highlighting them directly for designers.

Figure \ref{fig:global_results} shows close-up examples of the CCPT results.  Instead of relying on expert
demonstrations, the agent took many different paths:
in Figure \ref{fig:global_results}(a) the agent uses a missing
collision box in a tiny portion of the wall, thus arriving into the goal area by exploiting a slight slope of the wall of a pillar; in Figures \ref{fig:global_results}(b) and \ref{fig:global_results}(c) the agent exploits two oversights that allows it to jump over the wall and skip the main entrance. The first one uses an unintended prop near the wall while the second one exploits a climbable surface and precise double jumps; and in Figure \ref{fig:global_results}(d) the agent exploits a glitch that allows it to perform infinite jumps and to skip the wall like in the previous examples. This last issue is quite interesting as the agent has actually learned to exploit the glitch rather than using it at random.
    
    
Such examples can be also found in all of the four goal areas
tested in this paper, highlighting the good performance of our
algorithm. More high-resolution results are given in the
Supplementary Material.

\begingroup
\setlength{\tabcolsep}{10.5pt}
\begin{figure*}
    \begin{center}
    \scalebox{1.0}{
    \begin{tabular}{cccc}
        \includegraphics[width=0.21\textwidth]{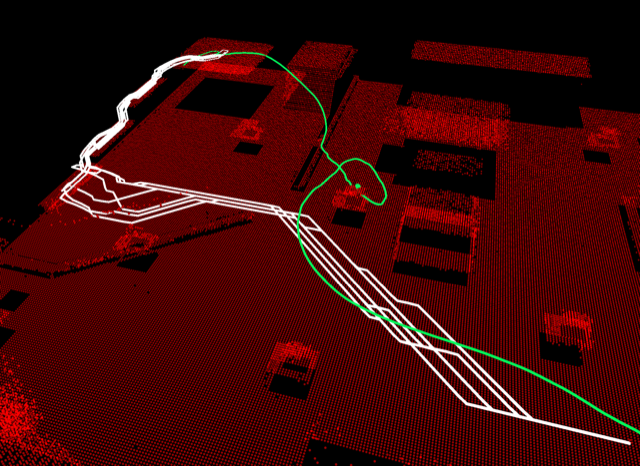} &
        \includegraphics[width=0.21\textwidth]{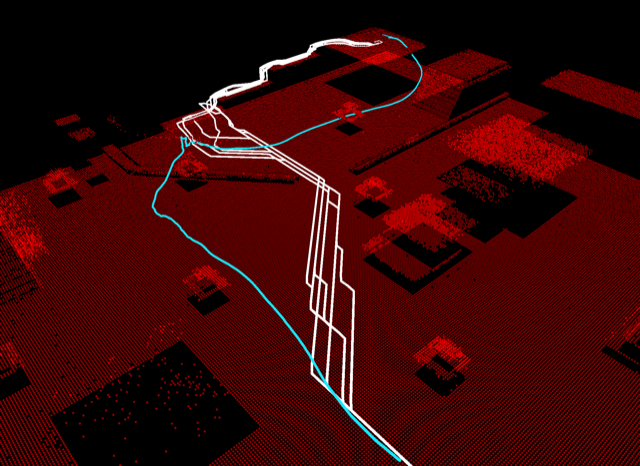} &
        \includegraphics[width=0.21\textwidth]{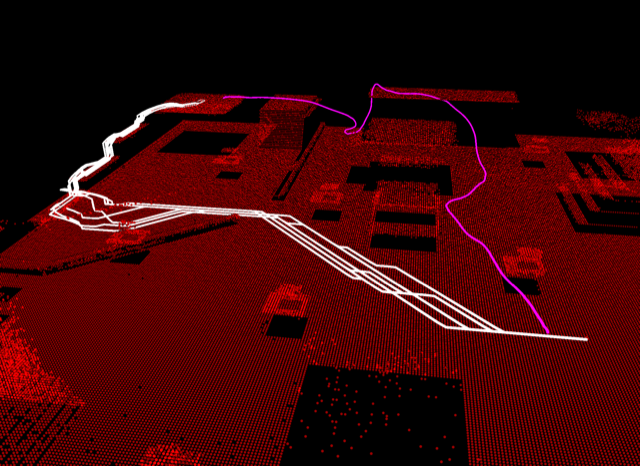} &
        \includegraphics[width=0.21\textwidth]{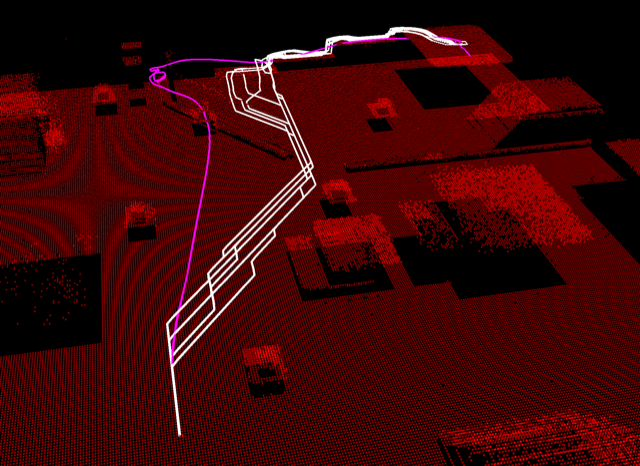}  \\
        (a) Missing collision box & (b) Exploiting props &
        (c) Using unintended path & (d) Exploiting glitches
    \end{tabular}
    }
    \end{center}
    \caption{
    Close-up images illustrating different
    bugs and gameplay issues found by CCPT. White trajectories indicate expert demonstrations, while the colored ones are those highlighted by CCPT as problematic.}
    \label{fig:global_results}
\end{figure*}
\endgroup

\renewcommand{\arraystretch}{1.3}
\begin{table}
  \begin{center}
  \scalebox{0.85}{
      \begin{tabular}{cccc}
        \toprule 
         & \textbf{Coverage} & \textbf{Bugs Found} & \textbf{Bugs Highlighted} \\
        \midrule
        CCPT & $1.91$ & $\textbf{13}$ & $\textbf{13}$ \\
        Linear Combination & $1.14$ & $7$ & $7$ \\
        Only Imitation & $0.84$ & $1$ & $0$ \\
        Only Curiosity & $\textbf{2.39}$ & $10$ & $3$ \\
      \end{tabular}
  }
  \caption{Quantitative results compared 
  to baselines: using a linear combination of curiosity and imitation; 
  using only imitation learning; and using only curiosity like
  \cite{improving}. Coverage
  is expressed in million of different states explored during training. The Bugs Found 
  column regards issues found by agents during  training, but not 
  necessarily identified as issues, while Bugs Highlighted are 
  issues found and identified as bugs by our model.}
  \label{tab:results} 
  \end{center}
\end{table}

\subsection{Evaluating the Reward Function}
\label{sec:abl_rew}
To evaluate the performance of our
\textit{exploration-conditioned intrinsic reward function}
we performed an ablation study using different fixed 
values of $\alpha$: $\alpha=0.5$ resulting in an average of $r_{\text{c}}$ and $r_{\text{i}}$,
$\alpha=0.0$ corresponding to only using imitation, and $\alpha=1.0$ corresponding to only using curiosity.


Table \ref{tab:results} shows the number of points in space covered for the four methods. As expected, agents trained with only imitation learning cover the smallest part of the environment, while
those trained with only curiosity cover the most points.  However, see that agents trained with our
algorithm achieve much better exploration when compared to only imitation learning and average of imitation and curiosity.  
This is an interesting finding: while they cover a very
large portion of the map, agents trained with our reward are still able
to follow the expert 
demonstrations. This enables us to filter 
and highlight broken trajectories as described in Section \ref{sec:alg}.

In Figure \ref{fig:abl_traj}(a) we give a top-down view of the
comparison of trajectories found by agents trained
with CCPT and those found with the average curiosity and imitation reward. As the plots show, the trajectories found with our reward cover a wider area, finding more differing and varying paths with respect to the baseline. 
We do not visualize plots for agents training using only the imitation module as they closely resemble the expert demonstrations.  We similarly omit plots for pure curiosity since, though it does find various way to arrive in the goal area, without the use of expert demonstrations there is no
way to tell if a trajectory is suspicious or similar to expected gameplay experience.

\begingroup
\setlength{\tabcolsep}{-1pt}
\begin{figure}
    \begin{center}
    \scalebox{1.0}{
    \begin{tabular}{cc}
        \includegraphics[width=0.25\textwidth]{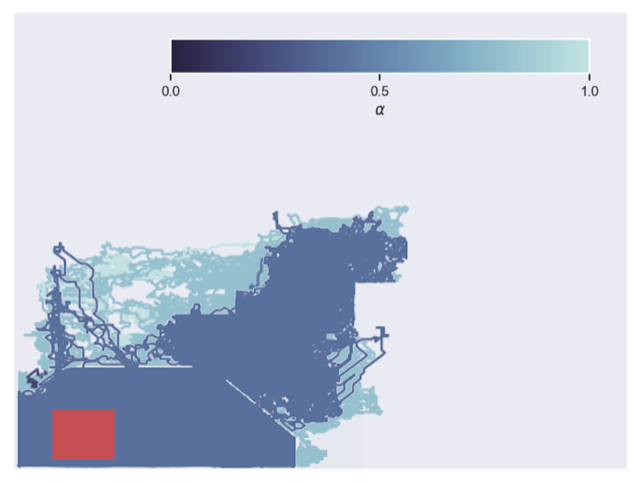} &  
        \includegraphics[width=0.25\textwidth]{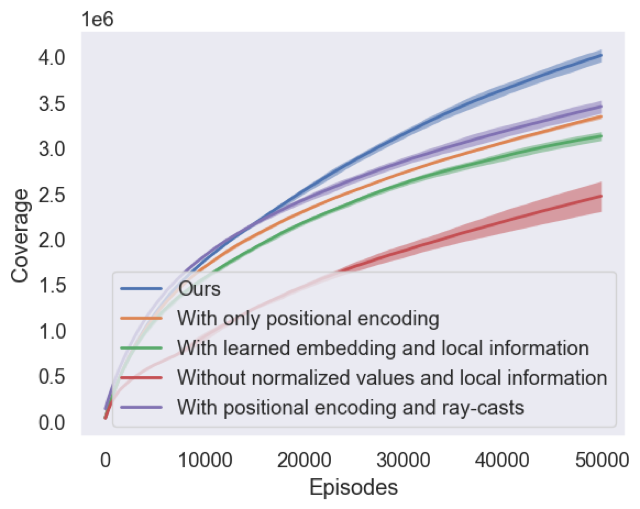} \\
        (a) & (b)
    \end{tabular}
    }
    \end{center}
    \caption{
        Results of ablation tests. Details in Sections 
        \ref{sec:abl_rew} and \ref{sec:abl_nav}.}
    \label{fig:abl_traj}
\end{figure}
\endgroup

\subsection{Ablation Study}
\label{sec:abl_nav}
We performed a series of ablation tests to better understand the performance of our models. In particular, we claim that
the combination of the semantic occupancy map and the positional embeddings described in Section \ref{sec:alg} is an efficient way to encode environment information when compared to standard navigation and exploration architectures.

In Figure \ref{fig:abl_traj}(b) we plot map coverage as a function of training steps for different configurations of the policy network
architecture. From the plot it is clear that the positional
embeddings used to encode the global positions provide an significant boost to the
performance compared to using only normalized values as in
\citep{improving, microsoft_navigation,ubisoftnavigation} or
even to learned embeddings. To the best of our knowledge, this is the first example of using such positional embeddings in deep reinforcement learning. Furthemore, the 
plot shows that the 3D semantic occupancy map and the
relative 3D convolutional network improve the results compared to using only global information and to a ray-casting baseline similar to \cite{improving}. The ray-casting approach uses 24 rays cast in various directions and at various heights. Each ray provides two values: the collision distance and the semantic value of the collided object.
From the plots it is clear that the combination of positional embeddings and semantic occupancy map that defines our full model clearly outperforms all other ablations.

\section{Conclusion and Discussion}
\label{sec:conclusion}
In this paper we introduced a novel reinforcement learning approach to automatically
play test complex 3D scenarios. Curiosity-Conditioned Proximal Trajectories (CCPT) enable developers and
designers to specify an area to test in the form of expert demonstrations.
CCPT uses a combination of imitation
learning and curiosity, driven by what we call an
\emph{exploration-conditioned intrinsic reward function}, to perform
exploration in the proximity of the demonstrated trajectories.
Our approach is not only able to find glitches and oversights,
but can also automatically identify and highlight trajectories containing potential issues. Our results show a high level of coverage
and bug discovery in our proposed navigation environment,
demonstrating how the particular combination of curiosity and
imitation works well for this purpose. 
We believe that this algorithm will be a useful tool for AAA game designers 
to automatically identify issues or potential exploits with less reliance on human testers.


A limitation of CCPT applied to automatic gameplay analysis is that we must perform one experiment for
each area we want to test. A possible solution would be to provide a large
set of demonstrations and to test different areas all at once. However, it
is known that GAN-based approaches are susceptible to mode collapse
when fit to big datasets. In this setting, the policy is prone
to imitate only a small subset of the example behaviors, thus
focusing only on one area. Another drawback is that CCPT
is not guaranteed to find all the bugs and issues in the
environment. In particular, those that are too far from the expert
demonstrations can be missed by our agents. One possible solution we
are experimenting with is to perform another CCPT iteration using
the problem trajectories found in the first iteration as expert
demonstrations: since they regard the same area, the policy
will not suffer from mode collapse while at the same time increasing exploration.


\bibliographystyle{named}
\bibliography{biblio}

\end{document}


\maketitle

\section{Hyperparameters}
The CCPT hyperparameters and their settings are shown in table \ref{tab:hyper}. Settings were chosen after a set of preliminary experiments made with different configurations. All training was performed deploying ten agents in parallel on the same machine with an NVIDIA RTX 2080 SUPER GPU with 8GB RAM and a AMD Ryzen 7 3700X 8-Core CPU.

\renewcommand{\arraystretch}{1.5}
\begin{table}[!h]
  \begin{center}
      \begin{tabular}{cc}
        \toprule 
        \multicolumn{2}{c}{\textbf{Navigation module}} \\
        \midrule 
        Learning rate $\alpha$ & $7e^{-5}$ \\
        Discount $\gamma$ & $0.90$ \\
        Entropy coefficient & $0.1$ \\
        \midrule 
        \multicolumn{2}{c}{\textbf{Imitation module}} \\
        \midrule 
        Learning rate $\alpha$ & $7e^{-5}$ \\
        Replay buffer size & $100000$ \\
        Batch size & $32$ \\
        Gradient penalty coefficient & $5.0$ \\
        \midrule 
        \multicolumn{2}{c}{\textbf{Curiosity module}} \\
        \midrule 
        Learning rate $\alpha$ & $7e^{-5}$ \\
        Batch size & $128$ \\
        \bottomrule
      \end{tabular}
  \end{center}
  \caption{Hyperparameters of CCPT. Values were chosen after a set of
    preliminary experiments made with different configurations.}
  \label{tab:hyper}
\end{table}

\section{Network architectures}
This section describes architectural details of each sub-module
discussed in section \ref{sec:method} of the main paper.

\begin{figure}
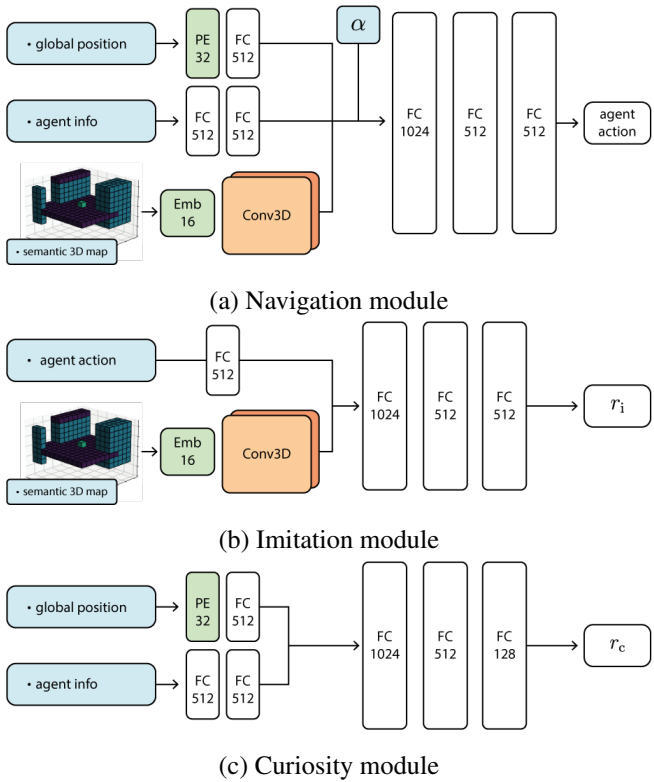

    \begin{center}
    \scalebox{1.0}{
    \begin{tabular}{c}
        \includegraphics[width=\columnwidth]{imgs/policy.png} \\
        (a) Navigation module \\
        \includegraphics[width=\columnwidth]{imgs/imitation.png} \\
        (b) Imitation module \\
        \includegraphics[width=\columnwidth]{imgs/curiosity.png} \\
        (c) Curiosity module
    \end{tabular}
    }
    \end{center}
    \caption{Overview of the module architectures used in this work.}
    \label{fig:nets_supplementary}
\end{figure}

\minisection{Navigation Module.} The navigation module, shown in figure 
\ref{fig:nets_supplementary}(a), takes three types
of inputs:
\begin{itemize}
    \item the $(X, Y, Z)$ integer global positions that are passed
    through the positional embedding with size 32 and a fully connected layer of size 512 with ReLu activation;
    
    \item the agent info vector (composed of values representing
    whether the agent is grounded, if is attached to a climbable surface, if it
    can perform a double jump, its velocity and direction) is passed
    through two fully connected layers of sizes 512 and ReLu activations;
    
    \item the local semantic 3D occupancy map of size 
    $21 \times 21 \times 21$.
    The map is a categorical discretization of 
    the space and elements around the agent,
    and each entry is the semantic integer value of the type of
    object at that position in $[0, 3]$. A value of 0 means an empty element, a 
    value of 1 means a solid shape (like solid ground, walls, obstacles, 
    moving elements), a value of 2 means a climbable surface and 
    a value of 3 means the agent itself. Each voxel covers a space
    of $\SI{1}{\m} \times \SI{1}{\m} \times \SI{1}{\m}$.
    The map is passed through a first
    embedding layer with size 16 and Tanh activation, then 
    through four 3D convolutional layers with 32, 32, 64, 64 channels, 
    each one with 
    kernel size $3 \times 3 \times 3$, stride 2 and ReLu activations. 
\end{itemize}
The outputs of the three branches are concatenated together with 
the $\alpha$ auxiliary input and
the resulting vector is passed through a feed forward network with 
three fully connected layers with sizes 1024, 512, and 512 and  ReLu 
activations. The final output is defined by another fully connected layer 
of size 10 and Softmax activation, representing the action probability
distribution.

Since we use the Actor-Critic PPO algorithm, we make use of a 
baseline network that has the same architecture as the policy network.

\minisection{Imitation Module.} The imitation module, shown in figure 
\ref{fig:nets_supplementary}(b), takes two types of inputs:
\begin{itemize}
    \item the one-hot encoded agent action that is passed through
    a fully connected layer of size 512 and ReLu activation;
    
    \item the local semantic 3D occupancy map. The shape of the map, as well
    as the architecture of the Conv3D network are the same of the 
    navigation module.
\end{itemize}
The outputs of the two branches are concatenated together and
the resulting vector is passed through a feed forward network with 
three fully connected layers of size 1024, 512 and 512 with ReLu 
activations. The final output layer is defined by a fully connected
layer with size 1 and Sigmoid activation representing the discriminator
probability $D(s,a)$.

\minisection{Curiosity Module.} The curiosity module, shown in figure 
\ref{fig:nets_supplementary}(c), takes two types of inputs:
\begin{itemize}
    \item the $(X, Y, Z)$ integer global positions that are passed
    through the positional embedding with size 32 and a fully connected
    layer of size 512 with ReLu activation;
    
    \item the agent information vector that is passed
    through two fully connected layers of sizes 512 and ReLu activations.
\end{itemize}
The outputs of the two branches are concatenated together and
the resulting vector is passed through a feed forward network with two fully connected layers of size 1024 and 512 with ReLu activations and a last fully connected layer of size 128 without activation.

\section{Additional experiments}
This section describes additional experiments on the VizDoom environment~\citep{wydmuch2018vizdoom, vizdoom} and also provides more high-resolution visualizations of agents in both the VizDoom environment and our 3D navigation environment described in the main paper.

\subsection{VizDoom Experiments}
In this section we investigate the robustness of CCPT 
applied in a real game environment different from our 3D navigation environment. To verify that CCPT can be a useful tool for game development, we applied it to the VizDoom environment, which is a semi-realistic 3D world based on the classic first-person shooter video game Doom~\citep{vizdoom}. The model architectures for all the modules are the
same of section \ref{sec:alg} 
of the main paper, but replacing the 3D semantic occupancy
map and the relative 3D convolutional network with the renderer depth buffer 
and a 2D convolutional network. The depth buffer has size $32 \times 32$ and
the Conv2D is composed of two 2D convolutional layers with 32 and 64 channels.
each one with kernel size $3 \times 3$, stride 2 and ReLu activation.
The action space consists of 6 different
actions: move forward, move left, move right, turn left, turn right and jump.

We tested our algorithm on the level shown in figure \ref{fig:vizglobal}(a).
The scene is approximately  $\SI{2800}{\m} \times \SI{2800}{\m} \times \SI{100}{\m}$ in size. Since we are interested in coverage testing, we removed all enemies and props. 
All hyperparameters used in this experiment are the same as those 
shown in table \ref{tab:hyper}.

\begingroup
\setlength{\tabcolsep}{10pt}
\begin{figure*}
    \begin{center}
    \scalebox{1.0}{
    \begin{tabular}{cc}
        \includegraphics[width=0.35\textwidth]{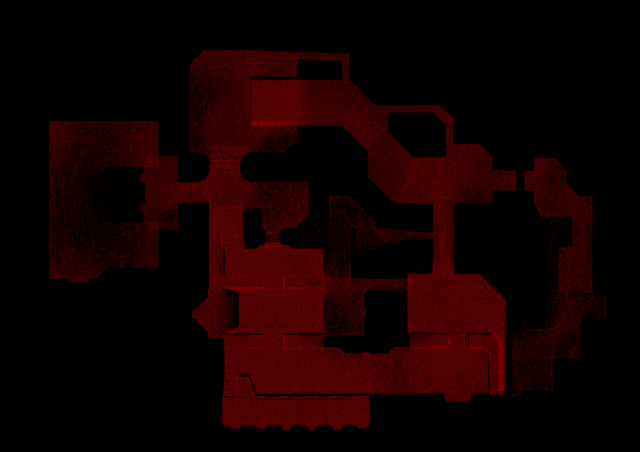} &
        \includegraphics[width=0.35\textwidth]{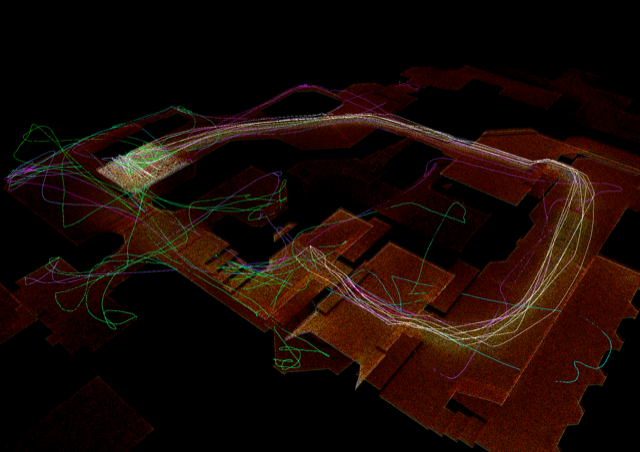}  \\
        (a) Tested level & (b) Global results
    \end{tabular}
    }
    \end{center}
    \caption{
        Results of the VizDoom experiment. (a) The
        explored level. (b) The 
        results of the experiment: trajectories in white are
        the expert demonstrations, while the colored ones are the results
        of CCPT.
    }
    \label{fig:vizglobal}
\end{figure*}
\endgroup

\begingroup
\setlength{\tabcolsep}{1pt}
\begin{figure*}
    \begin{center}
    \scalebox{1.0}{
    \begin{tabular}{cccc}
        \includegraphics[width=0.25\textwidth]{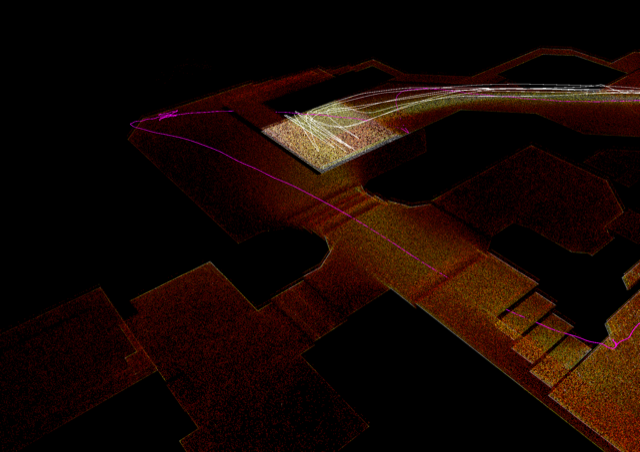} &
        \includegraphics[width=0.25\textwidth]{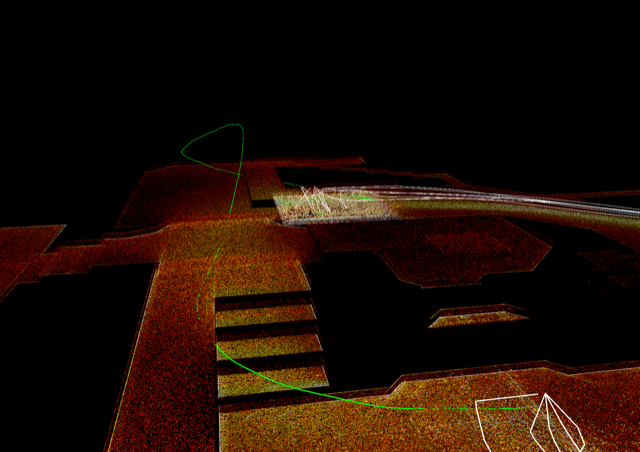} &
        \includegraphics[width=0.25\textwidth]{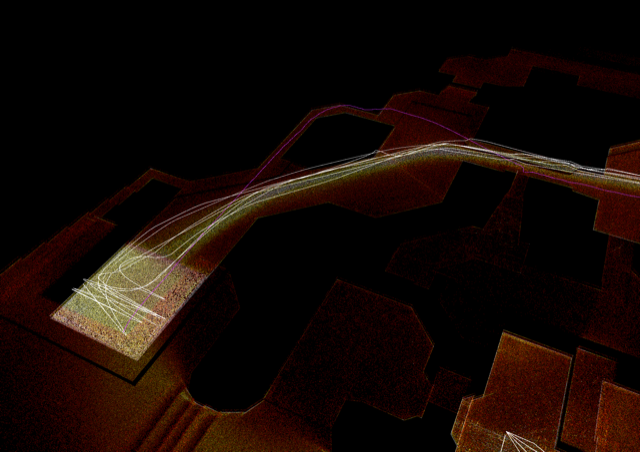} &
        \includegraphics[width=0.25\textwidth]{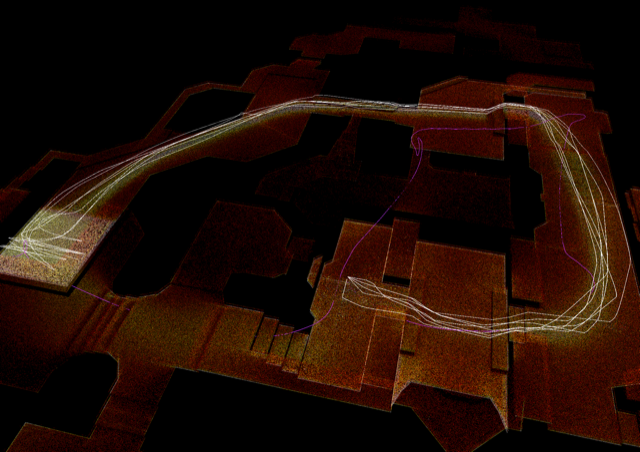}  \\
        (a) & (b) & (c) & (d)
    \end{tabular}
    }
    \end{center}
    \caption{
        Close-up images from the VizDoom experiment. Each image represents a different
        bug or issue found by CCPT. The white trajectories describe 
        the expert demonstrations, while the colored ones are those 
        highlighted by CCPT.
    }
    \label{fig:vizlocal}
\end{figure*}
\endgroup

For the experiment, we recorded expert demonstrations showing the intended path
for arriving at a specific goal area. We then let CCPT run and in the end we
visualize the most relevant trajectories highlighted by the algorithm.
Figure \ref{fig:vizglobal}(b) shows some qualitative results in which we can see how the algorithm is perfectly able to find and highlight different ways of arriving to the same 
goal area with respect to expert trajectories shown in white.
In figure \ref{fig:vizlocal} we give some close-up examples of specific findings:
\begin{itemize}
    \item figure \ref{fig:vizlocal}(a) shows how the agent can arrive at the goal area using a completely
    different path than the intended one;
    
    \item figure \ref{fig:vizlocal}(b) shows a trajectory similar to the one
    before, but using an elevated spot that was not meant to be reachable;
    
    \item figure \ref{fig:vizlocal}(c) shows a slight variation of expert
    demonstrations using a hidden path; and
    
    \item figure \ref{fig:vizlocal}(d) shows how the agent can use a complex
    path to skip part of the level and arrive in the goal location using
    a tiny gap between two walls.
\end{itemize}

\subsection{Additional visualizations}
Here we provide additional images of the results described in section
\ref{sec:exp} of the main paper. In figure \ref{fig:hr_global} we give visualizations of the results of the four goal area experiments in our 3D navigation environment.
Figure \ref{fig:hr_local} shows close-up examples of bugs found and 
highlighted by the algorithm.

\begingroup
\setlength{\tabcolsep}{1pt}
\begin{figure*}
    \begin{center}
    \scalebox{1.0}{
    \begin{tabular}{cccc}
        \includegraphics[width=0.25\textwidth]{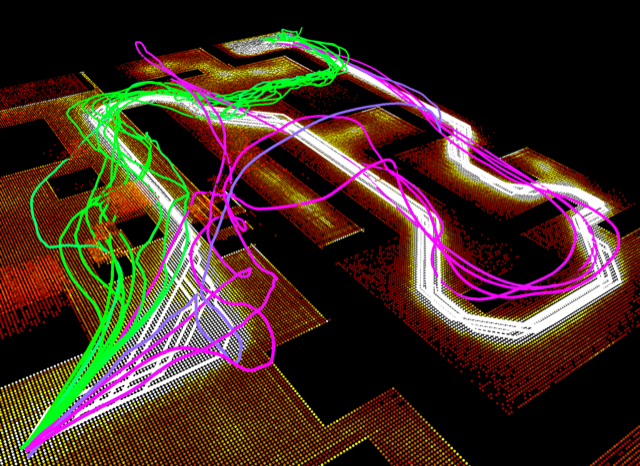} &
        \includegraphics[width=0.25\textwidth]{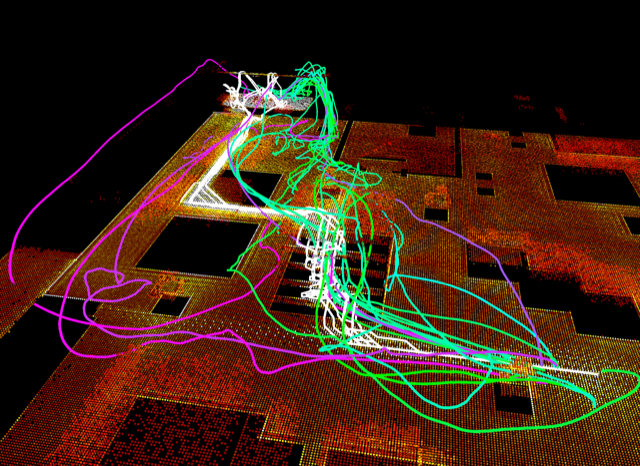} &
        \includegraphics[width=0.25\textwidth]{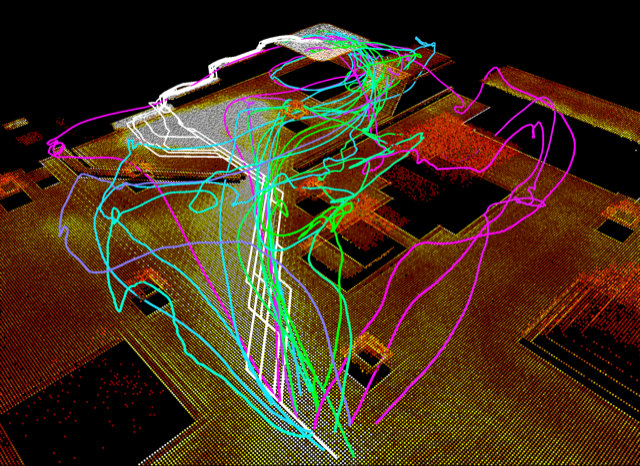} &
        \includegraphics[width=0.25\textwidth]{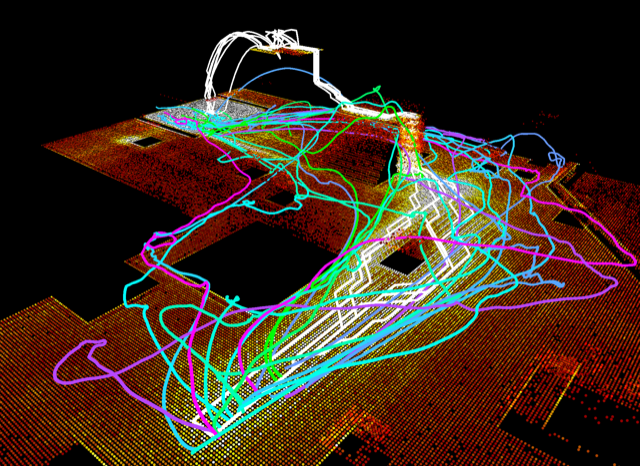}  \\
        (a) Area 1 & (b) Area 2 & (c) Area 3 & (d) Area 4 \\
    \end{tabular}
    }
    \end{center}
    \caption{
    Qualitative visualizations on four different tested areas. The white trajectories indicate the expert demonstrations, while the colored ones are those highlighted by CCPT.}
    \label{fig:hr_global}
\end{figure*}
\endgroup

\begingroup
\setlength{\tabcolsep}{1pt}
\begin{figure*}
    \begin{center}
    \scalebox{1.0}{
    \begin{tabular}{cccc}
        \includegraphics[width=0.25\textwidth]{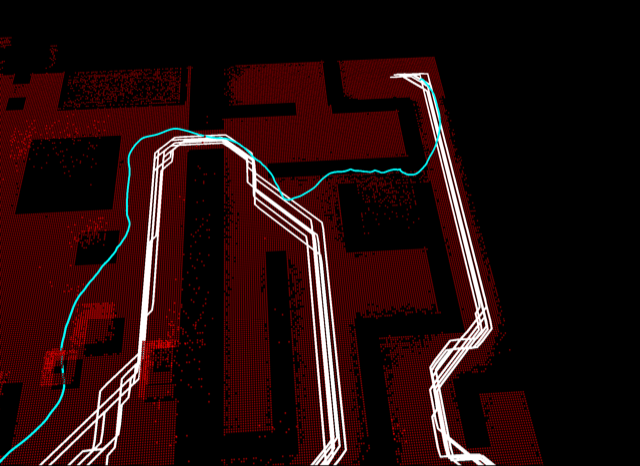} &
        \includegraphics[width=0.25\textwidth]{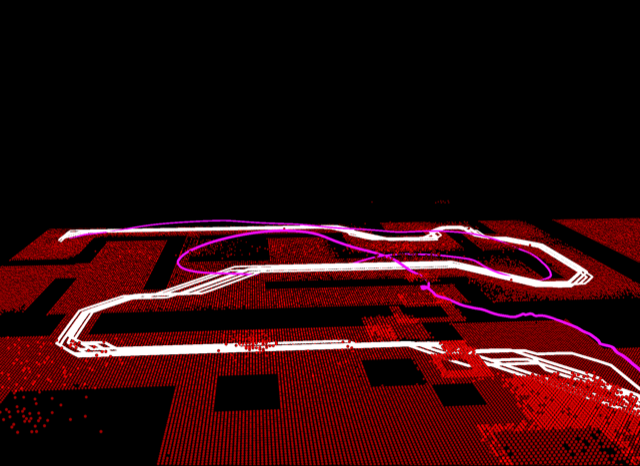} &
        \includegraphics[width=0.25\textwidth]{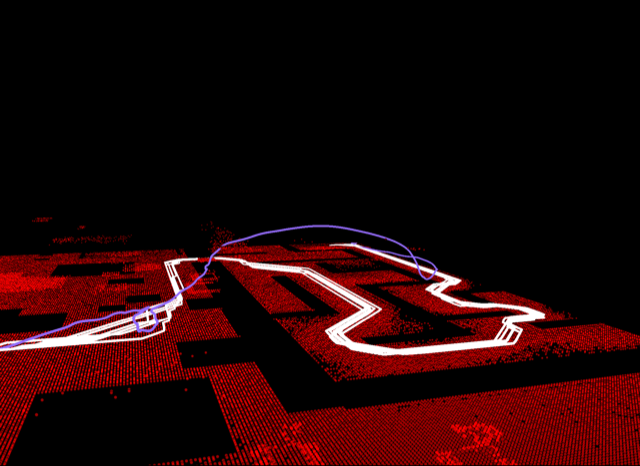} &
        \includegraphics[width=0.25\textwidth]{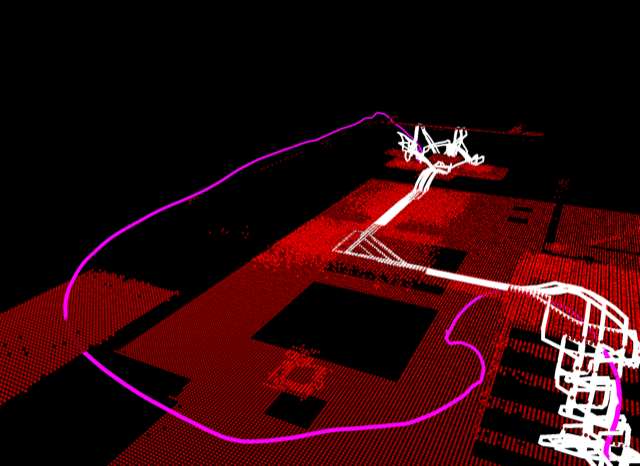}  \\
        (a) Missing collision box & (b) Skip main entrance & (c) Skip labyrinth & (d) Using unintended path \\
        \includegraphics[width=0.25\textwidth]{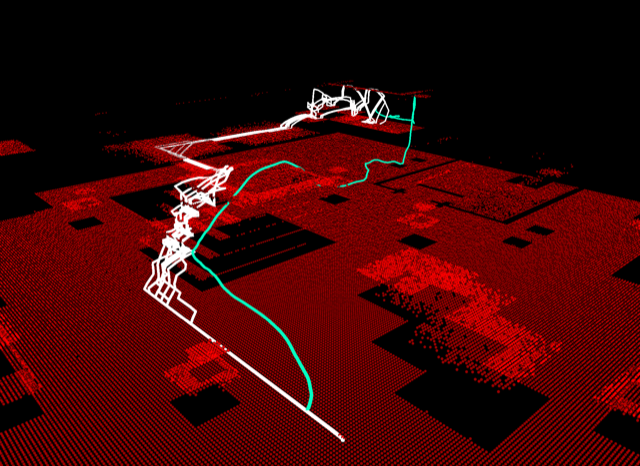} &
        \includegraphics[width=0.25\textwidth]{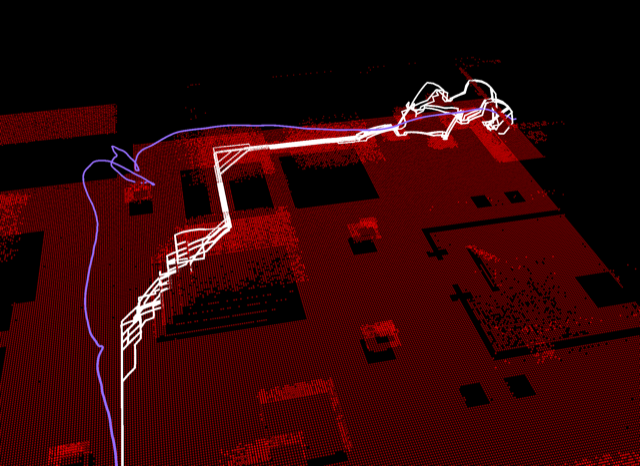} &
        \includegraphics[width=0.25\textwidth]{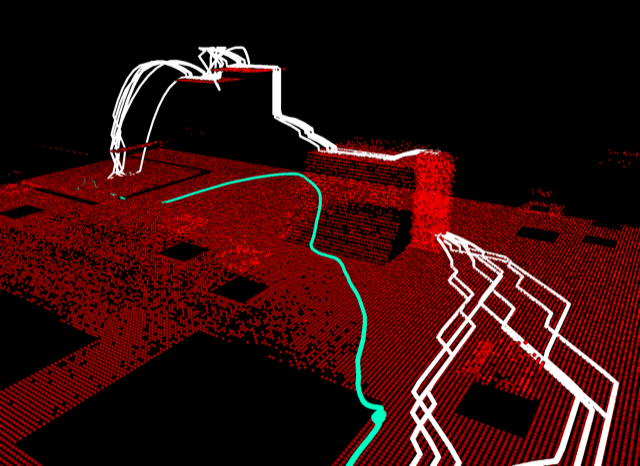} &
        \includegraphics[width=0.25\textwidth]{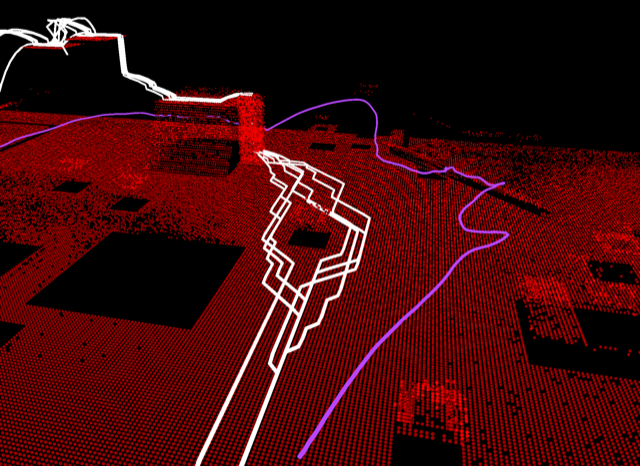}  \\
        (e) Using glitch & (f) Exploiting props & (g) Exploiting slope & (h) Using glitch \\
        \includegraphics[width=0.25\textwidth]{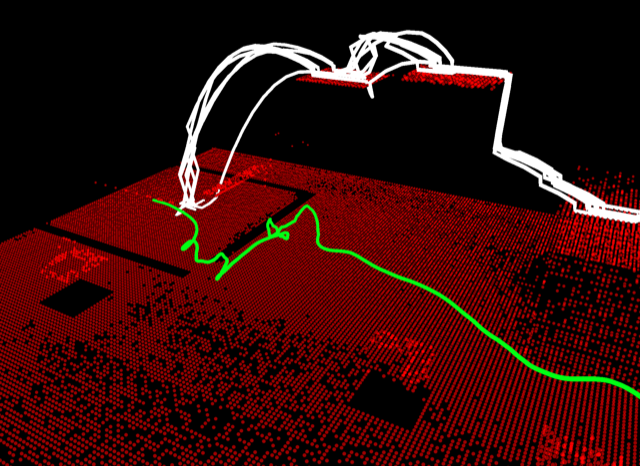} &
        \includegraphics[width=0.25\textwidth]{imgs/bug_1.png} &
        \includegraphics[width=0.25\textwidth]{imgs/bug_2.png} &
        \includegraphics[width=0.25\textwidth]{imgs/bug_3.png}  \\
        (i) Missing collision box & (l) Missing collision box & (m) Using glitch & (n) Exploiting props \\
        \multicolumn{4}{c}{\includegraphics[width=0.25\textwidth]{imgs/bug_4.png}} \\
        \multicolumn{4}{c}{(o) Using unintended path}\\
    \end{tabular}
    }
    \end{center}
    \caption{
    Close-up images of the four area experiments. Each image represents a different bug or issue found by CCPT. The white trajectories indicate expert demonstrations, while the colored ones are those highlighted by CCPT.}
    \label{fig:hr_local}
\end{figure*}
\endgroup

\bibliographystyle{named}
\bibliography{biblio}